\documentclass[sigconf]{acmart}

\usepackage{multirow}
\usepackage{microtype}
\usepackage{xspace}
\usepackage{balance}
\usepackage[ruled,vlined,linesnumbered]{algorithm2e}

\AtBeginDocument{%
  \providecommand\BibTeX{{%
    \normalfont B\kern-0.5em{\scshape i\kern-0.25em b}\kern-0.8em\TeX}}}

\copyrightyear{2022} 
\acmYear{2022} 
\setcopyright{acmcopyright}\acmConference[CIKM '22]{Proceedings of the 31st ACM International Conference on Information and Knowledge Management}{October 17--21, 2022}{Atlanta, GA, USA}
\acmBooktitle{Proceedings of the 31st ACM International Conference on Information and Knowledge Management (CIKM '22), October 17--21, 2022, Atlanta, GA, USA}
\acmPrice{15.00}
\acmDOI{10.1145/3511808.3557355}
\acmISBN{978-1-4503-9236-5/22/10}

\newcommand{\modelname}{CoLE\xspace}
\newcommand{\lmencoder}{N-BERT\xspace}
\newcommand{\kgeencoder}{N-Former\xspace}

\begin{document}

\title[Knowledge Graph Embedding via Co-distillation Learning]{I Know What You Do Not Know:\\ Knowledge Graph Embedding via Co-distillation Learning}

\author{Yang Liu}
\affiliation{
  \department{State Key Laboratory for Novel Software Technology}
  \institution{Nanjing University \country{China}}
}
\email{yliu20.nju@gmail.com}

\author{Zequn Sun}
\affiliation{
  \department{State Key Laboratory for Novel Software Technology}
  \institution{Nanjing University \country{China}}
}
\email{zqsun.nju@gmail.com}

\author{Guangyao Li}
\affiliation{
  \department{State Key Laboratory for Novel Software Technology}
  \institution{Nanjing University \country{China}}
}
\email{gyli.nju@gmail.com}

\author{Wei Hu}
\authornote{Wei Hu is the corresponding author.}
\affiliation{
    \department{State Key Laboratory for Novel Software Technology}
    \department{National Institute of Healthcare Data Science}
    \institution{Nanjing University \country{China}}
}
\email{whu@nju.edu.cn}

\renewcommand{\shortauthors}{Yang Liu, Zequn Sun, Guangyao Li, \& Wei Hu}

\begin{abstract}
Knowledge graph (KG) embedding seeks to learn vector representations for entities and relations.
Conventional models reason over graph structures, but they suffer from the issues of graph incompleteness and long-tail entities.
Recent studies have used pre-trained language models to learn embeddings based on the textual information of entities and relations, but they cannot take advantage of graph structures.
In the paper, we show empirically that these two kinds of features are complementary for KG embedding.
To this end, we propose CoLE, a \textbf{Co}-distillation \textbf{L}earning method for KG \textbf{E}mbedding that exploits the complementarity of graph structures and text information. 
Its graph embedding model employs Transformer to reconstruct the representation of an entity from its neighborhood subgraph.
Its text embedding model uses a pre-trained language model to generate entity representations from the soft prompts of their names, descriptions, and relational neighbors.
To let the two model promote each other,
we propose co-distillation learning that allows them to distill selective knowledge from each other's prediction logits.
In our co-distillation learning, each model serves as both a teacher and a student.
Experiments on benchmark datasets demonstrate that the two models outperform their related baselines, 
and the ensemble method CoLE with co-distillation learning advances the state-of-the-art of KG embedding.
\end{abstract}



\maketitle

\section{Introduction}
\label{chapter:intro}

A knowledge graph (KG) is a multi-relational graph in which each node represents an entity and the directed edge has a label indicating the specific relation between two entities.
An edge in KGs is a triplet in the form of (\textit{head entity, relation, tail entity}), such as (\textit{Kobe Bryant, profession, Athlete}).
KGs play an important role in a variety of knowledge-driven applications such as question answering and recommender systems \cite{KG_survey}.
However, KGs are typically incomplete in the real world \cite{KGCompleteness},
which affects the performance of downstream tasks.
Researchers propose the task of link prediction to predict and complete the missing edges using KG embeddings \cite{LP_survery}.
Existing KG embedding models \cite{KGE_survey} have primarily focused on exploring graph structures, including scoring the edge plausibility \cite{TransE,ConvE,RotatE}, reasoning over paths \cite{PTransE,RDF2Vec,RSN}, as well as convolution or aggregation over neighborhood subgraphs \cite{RGCN,CompGCN,hitter}.
We refer to these studies as structure-based models.
Learning from graph structures is indifferent to what the name of an entity or relation is, but suffers from the incompleteness and sparseness issues, making it difficult to predict triplets of long-tail entities with few edges.

In recent years, knowledge probing studies such as LAMA \cite{lama} have revealed that pre-trained language models (PLMs for short, such as BERT \cite{BERT}) have the ability to store some factual and commonsense knowledge obtained from large amounts of textual corpus,
encouraging increased interest in probing PLMs to complete KGs.
PLM-based KG embedding models convert a triplet to a natural language-style sequence by splicing the names of entities and relations, such as ``\textit{Kobe Bryant profession Athlete}''.
The sequence is then encoded by a PLM.
The output representations are used to predict or generate the masked entity \cite{kgt5}.
PLM-based KG embedding models do not suffer from the incompleteness issue since the PLM has been trained using external open-domain corpus.

\begin{figure}[!t]
\centering
\includegraphics[width=0.8\linewidth]{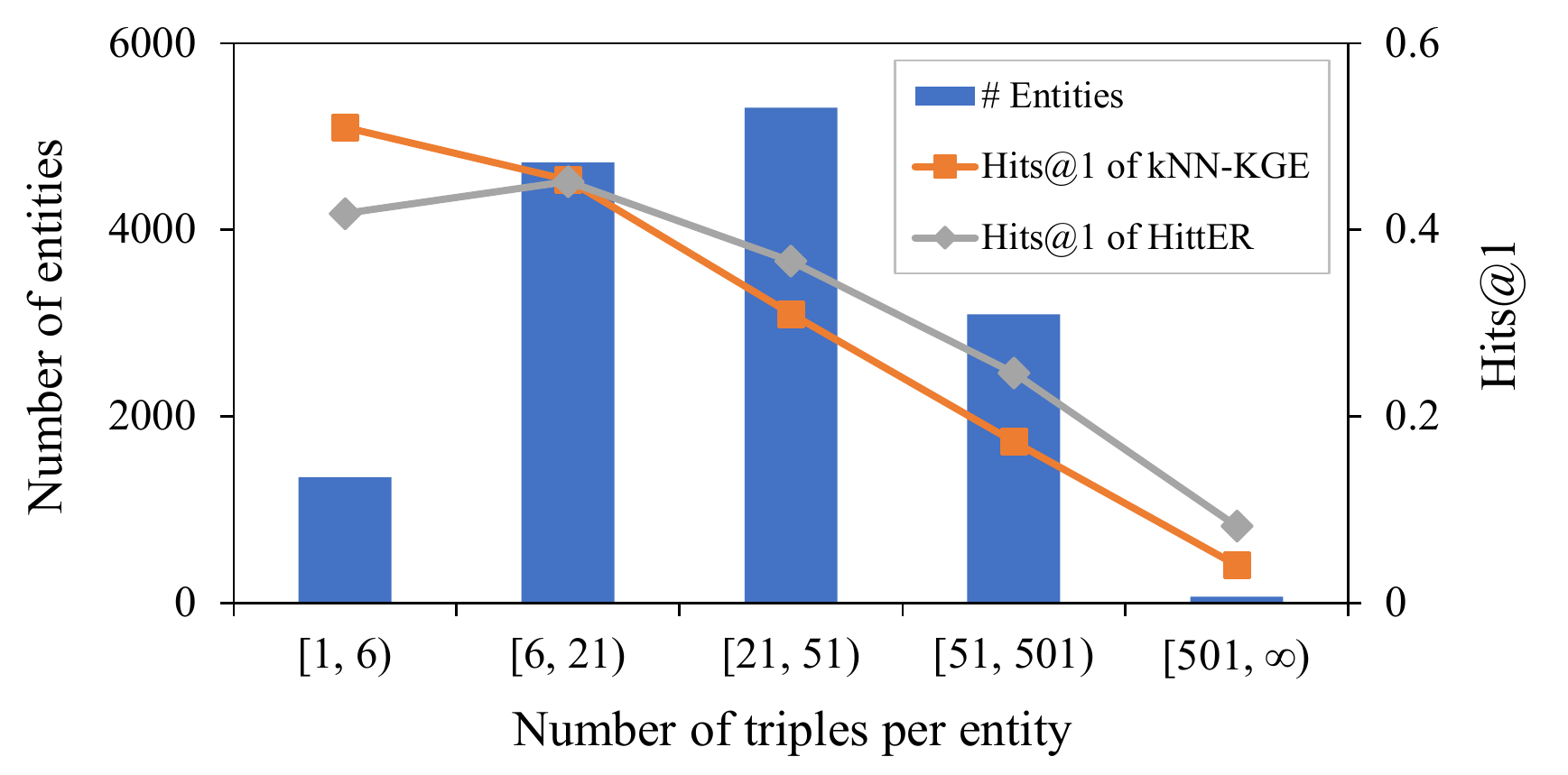}
\caption{Histogram: entities groups of FB15K-237 based on the number of triplets per entity. Line chart: the Hits@1 performance of the PLM-based model kNN-KGE \cite{knnkge} and structure-based model HittER \cite{hitter} in predicting the incomplete triplets of the entities in each group.}
\vspace{5pt}
\label{fig:long_tail}
\end{figure}

The two lines of KG embedding techniques are currently being studied separately.
But we would like to show that structure-based and PLM-based models have potential complementarity.
As a preliminary study, we partition the entities in the link prediction dataset FB15K-237 \cite{FB15K237} into several groups based on the number of edges (i.e., triplets) per entity,
and evaluate the performance of the PLM-based model kNN-KGE \cite{knnkge} and the structure-based model HittER \cite{hitter} in predicting the missing triplets of the entities in each group.
Figure \ref{fig:long_tail} exhibits the results.
kNN-KGE performs much better than HittER in link prediction of long-tail entities (see the results on the $[1,6)$ group),
because it can benefit from the external knowledge from PLMs.
HittER outperforms kNN-KGE for entities with rich edges.
This indicates that they are strongly complementary. 
Moreover, we can see that their results decline gradually in the groups of $[6, +\infty$).
The reason for the performance decline of HittER lies in that the rich edges of an entity typically involve multi-mapping (e.g., one-to-many and many-to-one) relations, which present the widely recognized challenge for structure-based models \cite{TransH,TransR}.
We also discover that the PLM-based model kNN-KGE may be affected by a similar issue.
Entities with rich edges are usually popular and can be found in many texts.
The noise, homonym, and ambiguity issues in such texts make it difficult for PLMs to recover the related content for predicting the missing triplets of a specific entity. 

\begin{figure}[!t]
\centering
\includegraphics[width=0.9\linewidth]{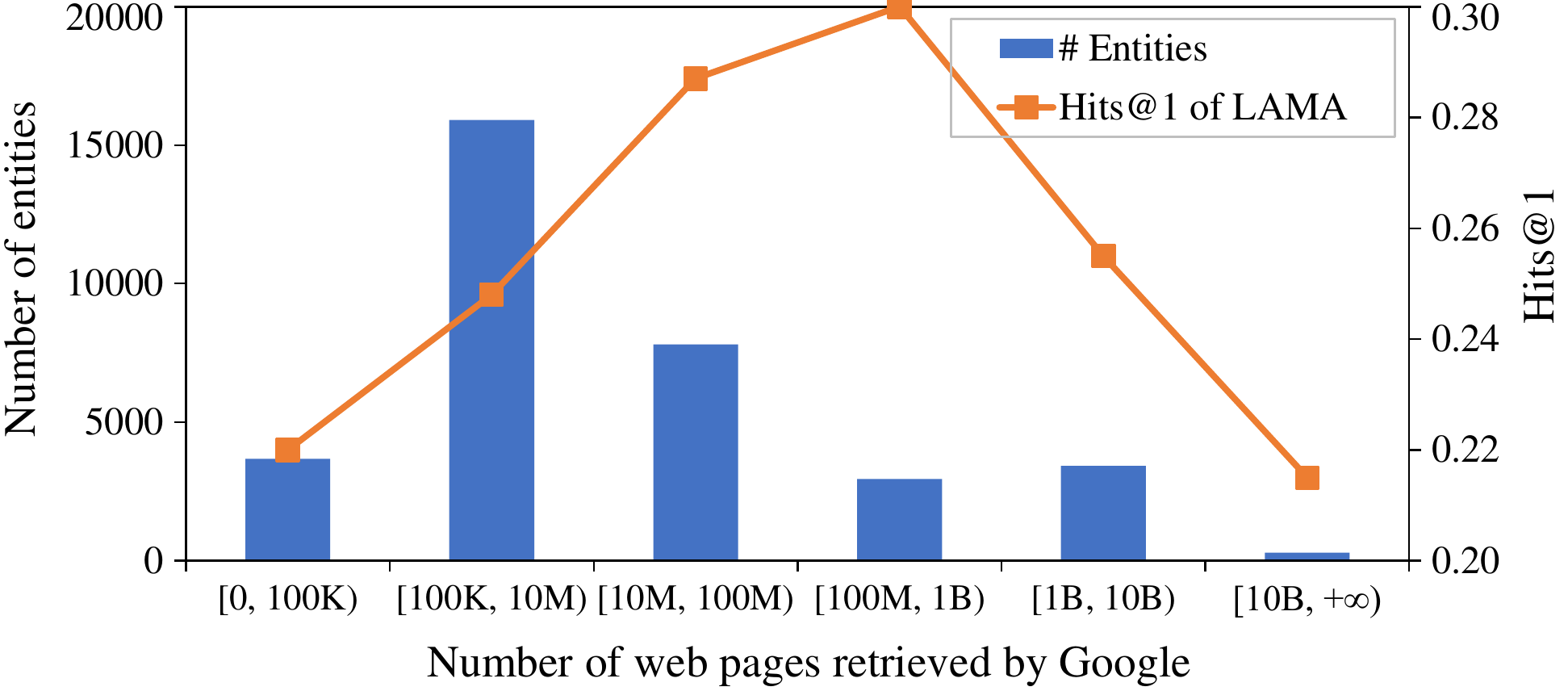}
\caption{Histogram: entities groups of T-Rex \cite{T-REx} and Google-RE \cite{lama} based on the number of web pages per entity retrieved by Google using the entity name as keywords. Line chart: the Hits@1 performance of LAMA \cite{lama} in predicting the missing triplets of the entities in each group.}
\label{fig:web_search}
\end{figure}

To investigate the aforementioned potential limitation of PLM-based models,
we use the entity names in T-REx \cite{T-REx} and part of Google-RE \cite{lama} as query keywords to Google Search, and
divide entities into several groups based on the number of retrieved web pages.
Figure~\ref{fig:web_search} depicts the groups along with the corresponding link prediction results of LAMA \cite{lama}.
We can see that LAMA fails to perform well in link prediction of popular entities (e.g., the $[1B,10B)$ group).
We find that the PLM-based models suffer from natural language ambiguity when confronted with some popular entities with names that are similar or even the same.
Another reason is that, although PLMs are trained using large amounts of textual corpus, it is difficult to retrieve the useful and to-the-point knowledge from PLMs to assist in the completion of a specific KG.

To take full advantage of the two types of models and resolve their limitations, 
in this paper we propose a novel approach, namely \modelname, for KG embedding via co-distillation learning between a structure-based model \kgeencoder and a PLM-based model \lmencoder.
The key idea is to let the two models selectively learn from and teach each other.
Specifically, \modelname consists of three components:
\begin{itemize}
    \item The structure-based model \kgeencoder employs Transformer \cite{Transformer} to reconstruct the missing entity of an incomplete triplet by leveraging the neighborhood subgraphs of the seen entity.
    Specifically, given an incomplete triplet $(h, r, ?)$, \kgeencoder first reconstructs the representation of $h$ from its neighbors. 
    This representation is then combined with the original representation of $h$ to further reconstruct the representation of the missing entity, $t$.
    Introducing neighborhood subgraphs in entity reconstruction can help resolve the issues of graph structure sparseness and long-tail entities.

    \item The PLM-based model \lmencoder is built upon BERT \cite{BERT} and seeks to generate the missing entity representation from a soft prompt that includes the description, neighbors, and names of the seen entity and relation.
    The description and neighbor information in the prompt can help retrieve the knowledge hidden in PLMs for a relevant specific entity.
    
    \item Our co-distillation learning method does not assume that one model is a teacher and the other is a student.
    We think that the two models are complementary in most cases.
    To let them benefit each other, we design two knowledge distillation (KD) objectives based on the decoupled prediction logits of the two models.
    The first objective seeks to transfer \kgeencoder's knowledge concerning the high-confidence predictions into \lmencoder, while the second is from \lmencoder to \kgeencoder.
    The prediction logits of a model are divided into two disjoint parts to calculate the two KD objectives, respectively, for selective knowledge transfer and avoiding negative transfer.
\end{itemize}

We conduct extensive experiments on benchmark datasets FB15K-237 \cite{FB15K237} and WN18RR \cite{ConvE} in three setting of the structure-based, PLM-based, and ensemble link prediction.
Results demonstrate that the two models, \kgeencoder and \lmencoder, achieve comparative and even better performance compared with existing related work, 
and the ensemble method CoLE with co-distillation learning advances the state-of-the-art of KG embedding, with a $0.294$ Hits@1 score on FB15K-237 and a $0.532$ Hits@1 score on WN18RR.

\section{Related Work}
\begin{figure*}[t]
\centering
\includegraphics[width=0.8\linewidth]{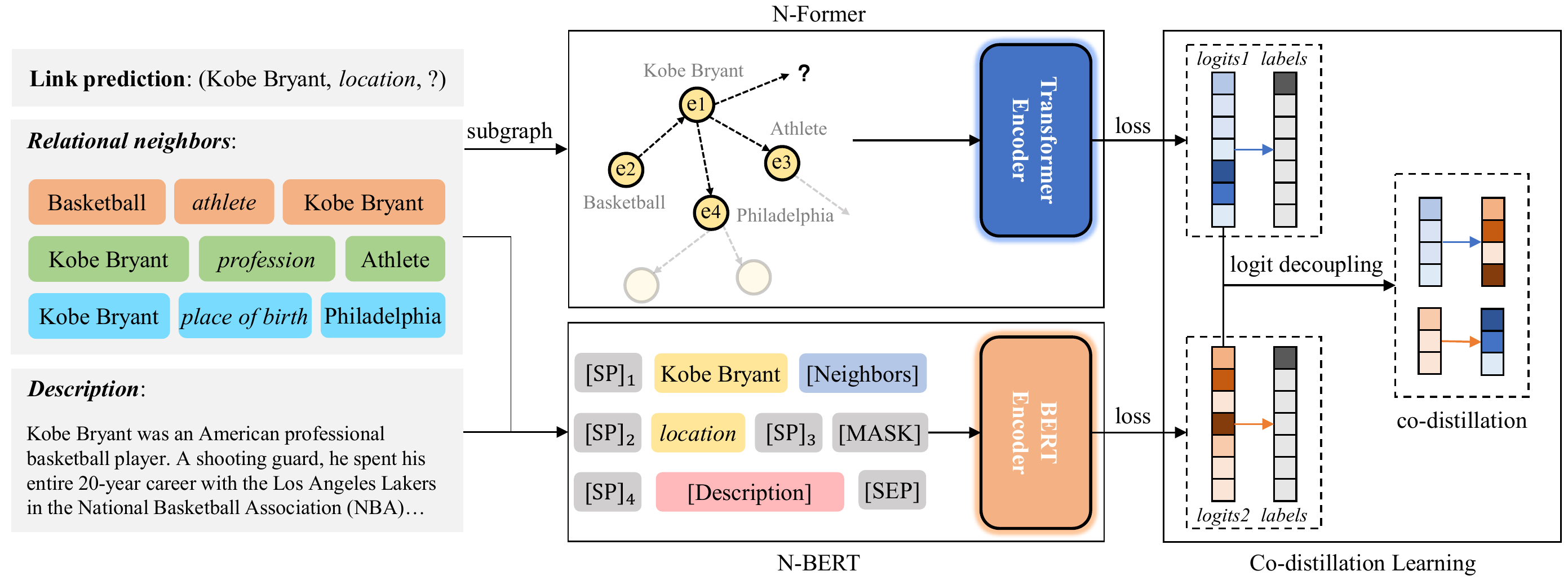}
\caption{Framework of the proposed co-distillation learning for KG embedding. It consists of three components: the structure-based model \kgeencoder, the PLM-based model \lmencoder, and the co-distillation method to let the two models teach each other.}
\label{fig:co_distillation}
\end{figure*}

\subsection{Structure-based KG Embedding}
We divide structure-based KG embedding models into three groups.
The first group contains the geometric models that interpret a relation in a triplet as a translation \cite{TransE,TransH,TransR} or rotation \cite{RotatE} from the head entity to the tail.
The second group uses bilinear functions \cite{ComplEx} or tensor decomposition methods to score triplets \cite{TuckER}.
The third group treats KG embedding as a deep learning task and explores various neural networks, including convolutional neural networks (CNNs) \cite{ConvE}, graph convolutional networks (GCNs) \cite{RGCN,CompGCN}, recurrent neural networks (RNNs) \cite{RSN}, hyperbolic neural networks \cite{ATTH} and Transformers \cite{CoKE,hitter}.
Readers can refer to the surveys \cite{KGE_survey,LP_survery} for an overview of the research progress.
Our \kgeencoder is a deep model based on Transformer \cite{Transformer}.
CoKE \cite{CoKE} does not consider the neighborhood subgraphs.
HittER \cite{hitter} uses Transformer in a GNN manner. 
It has an entity Transformer to aggregate each relational neighbor of an entity, and uses a context Transformer to aggregate the neighbor representations.
Despite the difference in network architectures between HittER and \kgeencoder,
their key difference lies in the different objectives.
HittER uses Transformers as a neighborhood aggregator to represent an entity (like CompGCN), 
while \kgeencoder is for entity reconstruction from incomplete triplets.

\subsection{PLM-based KG Embedding}
Unlike structure-based KG embedding models, some recent studies leverage PLMs to complete KGs by converting incomplete triplets to natural language queries.
LAMA \cite{lama} is a knowledge probing model which first reveals that PLMs can capture factual knowledge present in the training data and natural language queries structured as cloze statements are able to acquire such knowledge without fine-tuning the PLMs.
However, some strong restrictions, like manually constructed prompts and only predicting entities with single token names, hinder LAMA for the KG embedding task.
KG-BERT \cite{kg-bert} is the first model which applies PLMs to KG embedding, turning triplets into natural language sentences by simply concatenating entities' names and relations' names, then fine-tuning BERT for the sequence classification task.
Following KG-BERT, PKGC \cite{PKGC} leverages manual templates to construct coherent sentences that take full advantage of PLMs. 
It further adds entity definitions and attributes as support information.
KG-BERT and PKGC are both triplet classification models.
However, using the triplet classification models for link prediction is very time-consuming.
They usually assume all entities appearing in KGs are candidates, so they need numerous inference steps for one incomplete triplet, which is impractical when KGs are huge.
To predict the missing entities in one inference step, MLMLM \cite{mlmlm} adds more than one [MASK] token in the prompts to predict entities with multi-token names, while kNN-KGE \cite{knnkge} utilizes PLMs to learn an initial representation for each entity from its description.

Please note that our work is different from the studies injecting knowledge into PLMs to improve natural language processing (NLP) tasks.
We focus on the knowledge transfer and mutual enhancement between structure- and PLM-based models.
Our \lmencoder is a knowledge probing model and also differs from text-enhanced embedding studies \cite{StAR,KG-PubMedBERT} that integrate the text and structure features to represent entities.
\section{Approach}
This section introduces the proposed approach, \modelname.

\subsection{Notations}
A KG is denoted as a three-tuple $(\mathcal{E}, \mathcal{R}, \mathcal{T})$, 
where $\mathcal{E}$ is the set of entities, 
$\mathcal{R}$ is the set of relations,
and $\mathcal{T}$ is the set of triplets.
In a KG, an entity or relation is typically represented with a Uniform Resource Identifier (URI).
For example, the URI of Kobe Bryant in Freebase is \texttt{/m/01kmd4},
which, however, is not human-readable.
In our approach, we assume each entity $e\in \mathcal{E}$ and relation $r\in \mathcal{R}$ to have a human-readable literal name like ``Kobe Bryant'', which is denoted by $N_e$ and $N_r$, respectively.
Our approach \modelname also leverages the textual descriptions of entities to generate prompt templates.
The description of an entity $e$ is denoted by $D_e$.
We use $\mathbf{E}$ to denote an embedding. 
For example, $\mathbf{E}_e$ denotes the embedding of entity $e$,
and $\mathbf{E}_r$ denotes the embedding of relation $r$.

\subsection{Framework Overview}
Figure~\ref{fig:co_distillation} shows the overall framework of \modelname.
The objective is to predict the missing entity in an incomplete triplet, such as (Kobe Bryant, \textit{location}, ?) in FB15K-237.
The available information used to support this prediction includes
the relational neighbors of Kobe Bryant and its textual description. 
The proposed \kgeencoder first takes as input the subgraph to reconstruct a representation for Kobe Bryant.
The representation together with the initial embeddings of Kobe Bryant and the relation \textit{location} are then fed into \kgeencoder to reconstruct a representation for the missing entity denoted as a special placeholder ``[MASK]'' in the input.
The output ``[MASK]'' representation is used to compute the prediction logits.
Furthermore, the PLM-based model \lmencoder constructs a prompt from the descriptions, neighbors and names of entities as the input of BERT.
The missing entity is also replaced with the placeholder ``[MASK]''.
The output representation of BERT for ``[MASK]'' is used to predict the missing entity.
We assume that the two models are complementary, and that they should teach each other what they are good at.
We propose a decoupled knowledge co-distillation method.
It first divides the logits into two parts, one concerning the ``high-confidence'' predictions of \kgeencoder and the other for \lmencoder.
Then, it enables bidirectional knowledge transfer by minimizing the KL divergence of the partial logits from the two models.

\subsection{Neighborhood-aware Transformer}
\begin{figure}[!t]
\centering
\includegraphics[width=\linewidth]{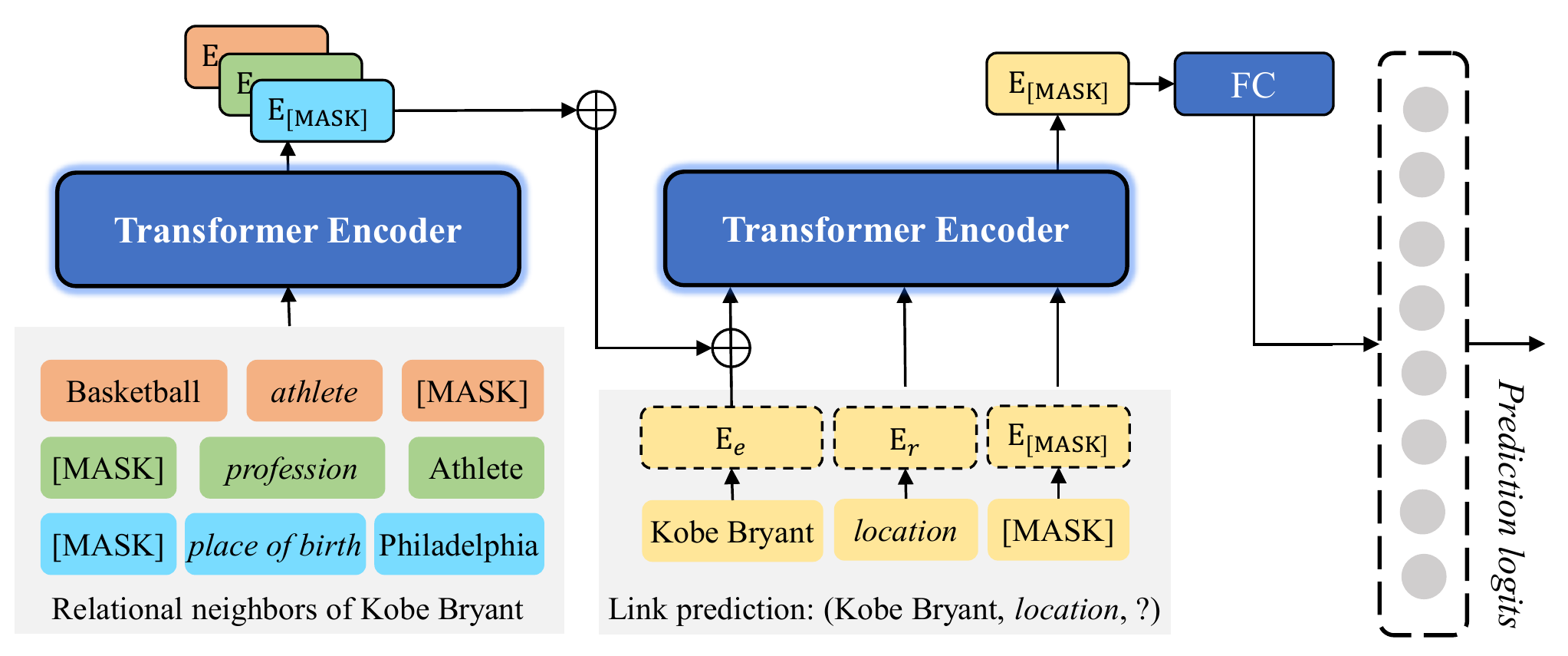}
\caption{Architecture of the proposed \kgeencoder.}
\label{fig:struc_former}
\end{figure}

Figure~\ref{fig:struc_former} shows the architecture of the proposed \kgeencoder with Transformer \cite{Transformer} as the backbone.
\kgeencoder is built on the idea of recursive entity reconstruction,
which enables the model to see more for prediction while remaining robust to graph incompleteness and sparseness.
We introduce the technical details below.

\subsubsection{Entity Reconstruction from a Triplet}
Given a triplet $(h, r, t)$,
we use the placeholder ``[MASK]'' to take the place of entity $t$, 
and our objective is to reconstruct the representation of $t$ from the incomplete triplet $(h, r, \text{[MASK]})$.
We feed the input embeddings of $(h, r, \text{[MASK]})$, which are randomly initialized, into Transformer and get the output representation for [MASK]:
\begin{equation}
\label{eq:transformer_base}
\mathbf{E}_{\text{[MASK]}}^{n}=\texttt{Transformer}(\mathbf{E}_{h}^{0},\mathbf{E}_{r}^{0},\mathbf{E}_{\text{[MASK]}}^{0}),
\end{equation}
where $n$ denotes the number of self-attention layers in Transformer.
With the help of self-attention, 
the output representation of [MASK] can capture the information from the entity $h$ and relation $r$ \cite{CoKE}.
We use it as the reconstruction representation of entity $t$, 
i.e., $\mathbf{E}_t =\mathbf{E}_{\text{[MASK]}}^{n}$,
and define the prediction logits $P_{t}$ as 
\begin{equation}
\label{eq:triple_predict}
\mathbf{P}_t=\texttt{softmax}\big(\mathbf{E}_{\text{ENT}}\cdot \texttt{MLP}(\mathbf{E}_t)\big),
\end{equation}
where $\mathbf{E}_{\text{ENT}}$ is the embedding matrix for all entities,
and $\texttt{MLP}()$ denotes a multi-layer perceptron for representation transformation.
We can then calculate the cross-entropy loss between the prediction logits $\mathbf{P}_t$ and the corresponding labels $\mathbf{L}_{t}$ as follows:
\begin{equation}
\label{eq:loss_triple}
\mathcal{L}_{\mathrm{triplet}}(t)=\texttt{CrossEntropy}(\mathbf{P}_t, \mathbf{L}_{t}),
\end{equation}
where $\texttt{CrossEntropy}(\mathbf{x},\mathbf{y})=-\sum\nolimits_j \mathbf{x}_j \log \mathbf{y}_j$, which computes the cross-entropy loss between two vectors.

\subsubsection{Entity Reconstruction from Neighborhood}
\label{method:kge_neighbor}

The entity reconstruction from a triplet described above can be used in a recursive manner to consider the neighborhood subgraphs.
For example, for the incomplete triplet $(h, r, \text{[MASK]})$,
in addition to the prediction loss in Eq.~(\ref{eq:triple_predict}) that only considers the input embeddings of $h$ and $r$,
we think that the embedding of $h$ can also benefit from its representation reconstructed from its relational neighbors,
which also lets the model see more to help with the prediction of long-tail entities.
Let $\texttt{Neighbor}(h)=\{(h',r')\,|\,(h',r',h)\in \mathcal{T}\}$ denote the set of relational neighbors of entity $h$.
Please note that we add a reverse triplet $(t, r^-, h)$ for each triplet $(h, r, t)$ in the KG, such that we only need to consider the incoming edges of an entity, i.e., the triplets whose tails are this entity.
We use each relational neighbor of $h$ to reconstruct a representation and sum up all these representations as the embedding of $h$, denoted by $\mathbf{E}_{Neighbor}^S$:
\begin{equation}
\label{eq:transformer_neighbor}
\resizebox{.92\columnwidth}{!}{$
\mathbf{E}_{Neighbor}^{S}=\sum_{(h',r')\in \texttt{Neighbor}(h)}\texttt{Transformer}(\mathbf{E}_{h'}^{0},\mathbf{E}_{r'}^{0},\mathbf{E}_{\text{[MASK]}}^{0}),
$}
\end{equation}
where the Transformer is the same one as in Eq.~(\ref{eq:transformer_base}).
To let the neighborhood information of $h$ be used in link prediction, we further expand Eq.~(\ref{eq:transformer_base}) by combining the input embedding of $h$ $\mathbf{E}_{h}^{0}$ with the embedding reconstructed from relational neighbors $\mathbf{E}_{Neighbor}^{S}$:
\begin{equation}
\label{eq:transformer_output}
\mathbf{E}_{\text{[MASK]}}^{S}=\texttt{Transformer}(\texttt{mean}(\mathbf{E}_{h}^{0},\mathbf{E}_{Neighbor}^{S}),\mathbf{E}_{r}^{0},\mathbf{E}_{\text{[MASK]}}^{0}),
\end{equation}
where $\texttt{mean}()$ returns the average of vectors.
Given the output representation, we can compute the prediction logits $\mathbf{P}_t^{S}$ in the same way as Eq.~(\ref{eq:triple_predict}).
The overall loss of \kgeencoder is defined as
\begin{equation}
\label{eq:loss_structure}
\mathcal{L}_{\mathrm{structure}}=\sum_{(h,r,t)\in \mathcal{T}}\big(\mathcal{L}_{\mathrm{triplet}}(t) + \texttt{CrossEntropy}(\mathbf{P}_t^{S}, \mathbf{L}_{t})\big).
\end{equation}

\begin{figure*}[!t]
\centering
\includegraphics[width=0.9\linewidth]{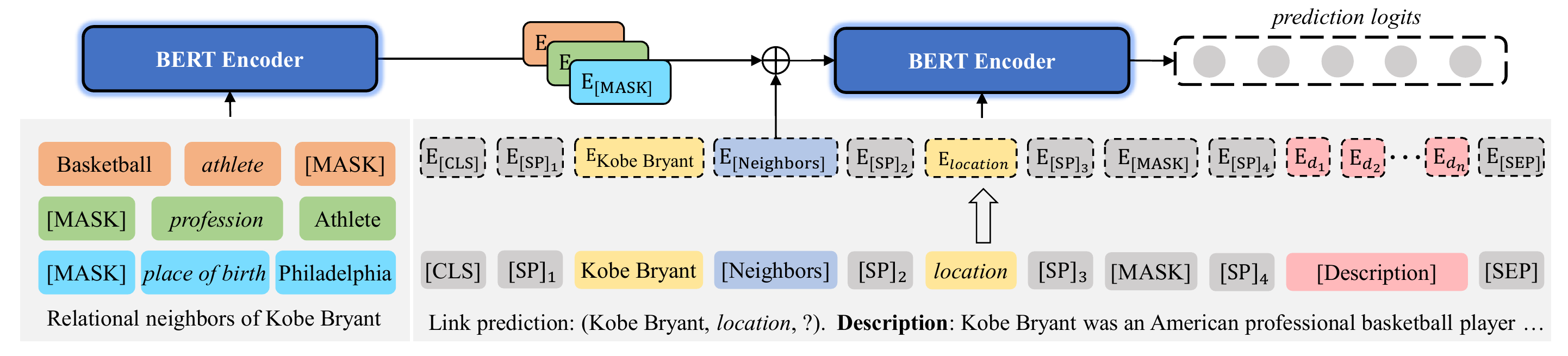}
\caption{Architecture of the proposed \lmencoder.}
\label{fig:text_former}
\end{figure*}

\subsection{Neighborhood-aware BERT}

\begin{table}
\caption{Prompt templates used in \lmencoder.}
\label{tab:prompt}
\centering
\resizebox{\linewidth}{!}{
    \begin{tabular}{ll}
        \toprule
        \multirow{2}{1.2cm}{Triplet prompt} & $\texttt{prompt}_T(h,r) =$ [CLS] $N_\text{h}$ [SEP] $N_\text{r}$ [SEP] [MASK] [SEP] $D_\text{h}$ [SEP] \\
        & $\texttt{prompt}_T(r,t) =$ [CLS] [MASK] [SEP] $N_\text{r}$ [SEP] $N_\text{t}$ [SEP] $D_\text{t}$ [SEP] \\
        \midrule
        \multirow{2}{1.2cm}{Relational prompt} & $\texttt{prompt}_R(h,r) =$ [CLS] [SP]$_1^{r}$ $N_\text{h}$ [SP]$_2^{r}$ $N_\text{r}$ [SP]$_3^{r}$ [MASK] [SP]$_4^{r}$ $D_\text{h}$ [SEP] \\
        & $\texttt{prompt}_R(r,t) =$ [CLS] [SP]$_1^{r}$ [MASK] [SP]$_2^{r}$ $N_\text{r}$ [SP]$_3^{r}$ $N_\text{t}$ [SP]$_4^{r}$ $D_\text{t}$ [SEP] \\
        \midrule
        \multirow{2}{1.2cm}{Neighbor prompt} & $\texttt{prompt}_N(h,r) =$ [CLS] [SP]$_1^{r}$ $N_\text{h}$ [Neighbors] [SP]$_2^{r}$ $N_\text{r}$ [SP]$_3^{r}$ [MASK] [SP]$_4^{r}$ $D_\text{h}$ [SEP] \\
        & $\texttt{prompt}_N(r,t) =$ [CLS] [SP]$_1^{r}$ [MASK] [SP]$_2^{r}$ $N_\text{r}$ [SP]$_3^{r}$ $N_\text{t}$ [Neighbors] [SP]$_4^{r}$ $D_\text{t}$ [SEP] \\
        \bottomrule
    \end{tabular}}
\end{table}

It has been widely acknowledged that PLMs like BERT \cite{BERT} capture some structured knowledge \cite{lama}, which can be leveraged for link prediction in KGs.
PLMs are trained with a large amount of open-domain corpora, but link prediction is about a specific entity in a KG.
The key challenge for PLM-based link prediction lies in how to retrieve the relevant knowledge from PLMs.
Given an incomplete triplet $(h, r, \text{[MASK]})$,
a typical solution is to introduce the textual description of the entity for constructing a sentence-like prompt \cite{knnkge} as \textit{triplet prompt} shown in Table~\ref{tab:prompt}, 
where [CLS], [SEP] are the special tokens used to separate the different parts of the prompt,
and the missing entity is replaced with a placeholder [MASK].
It is worth noting that we do not introduce inverse relations to convert an incomplete triplet $(\text{[MASK]}, r, t)$ as $(t, r^-, \text{[MASK]})$. 
This is because it is a non-trivial task to get the name of a reverse relation $r^-$ given the relation $r$.
For an incomplete triplet $(h, r, \text{[MASK]})$, its prompt sentence is denoted by $\texttt{prompt}(h, r)$.
For an incomplete triplet $(\text{[MASK]}, r, t)$, its prompt sentence is $\texttt{prompt}(r, t)$.
The prompt sentences would be fed into a PLM (BERT, in our work) to get the representation of [MASK]:
\begin{equation}
    \label{eq:bert_base}
    \mathbf{E}_{\text{[MASK]}} = \texttt{BERT}(\texttt{prompt}_T(h,r)),
\end{equation}
and we use it as the representation of $t$.

To get the predicted entities in one inference step, we need to add the names of entities as new tokens into the vocabularies of PLMs.
But the random initial embeddings for these new tokens perform poorly, as they do not own any external knowledge from open-domain corpora.
Following \cite{knnkge}, we use the embeddings learned from the description prompts as initial representations for entities:
\begin{equation}
    \mathbf{E}_{\text{h}}^{init} = \texttt{BERT}(\text{``}The\ description\ of\ \text{[MASK]}\ is\ D_h.\text{''}),
\end{equation}
where $D_h$ is the textual description of the entity $h$.
To make our prompts more expressive and take full advantage of the PLMs, we introduce two types of additional textual information in prompts.

\subsubsection{Soft Prompts}
Given that PLMs are designed for NLP, it is convincing that more coherent prompts would utilize the knowledge in PLMs better, while our prompts concatenated by [SEP] are obviously unexpressive.
Some works \cite{lama, PKGC} design manual templates to make the prompts more coherent, which is impractical for KGs with numerous relations.
On the contrary, we utilize some adjustable soft prompts to make our prompt more expressive.
Inspired by PKGC \cite{PKGC}, we replace the special [SEP] tokens between other tokens with relation-aware soft prompts to get a new kind of prompts, which is named \textit{relational prompt} in Table~\ref{tab:prompt}, where $\text{[SP]}_i^{r} (i = 1, 2, 3, 4)$ denotes the $i$-th soft prompt for the relation $r$ and $i$ indicates the position where the corresponding soft prompt is to be inserted.
In implementation, $\text{[SP]}_i^{r}$ is a special token added into the vocabulary and related to the relation $r$, and its representation is randomly initialized.

\subsubsection{Neighborhood Prompts}
Recent studies \cite{mlmlm, PKGC, knnkge} have shown that more support textual information leads to more promising performance, and the descriptions and attributes of entities have been widely used. 
However, the relational knowledge in KGs has not been explored for prompts, which may potentially facilitate the PLMs for link prediction.
To take full advantage of the neighboring triplets in the KGs, we introduce the contextual embeddings of entities as additional support information, which can also be regraded as entity-aware soft prompts.
Let $\texttt{InNeighbor}(h)=\{(h',r')\,|\,(h',r',h)\in \mathcal{T}\}$ and $\texttt{OutNeighbor}(h)= \{(r',t')\,|\, (h,r',t')\in \mathcal{T} \}$ denote the sets of relational neighbors of entity $h$.
Similar to Section~\ref{method:kge_neighbor}, we use neighbors of $h$ to reconstruct a representation and sum up all these representations as the embedding of $h$, denoted by $\mathbf{E}_{Neighbor}^{T}$:
\begin{equation}
\begin{aligned}
    \mathbf{E}_{Neighbor}^{T} &= \sum_{(h',r')\in \texttt{InNeighbor}(h)}\texttt{BERT}(\texttt{prompt}_R(h',r')) \\
                   &\quad + \sum_{(r',t')\in \texttt{OutNeighbor}(h)}\texttt{BERT}(\texttt{prompt}_R(r',t')),
\end{aligned}
\end{equation}
where the BERT is the same one as in Eq.~(\ref{eq:bert_base}).
After adding the neighbor information, our neighborhood prompt is named \textit{neighbor prompt} in Table~\ref{tab:prompt},
where $\text{[Neighbor]} = \mathbf{E}_{Neighbor}^{T}$ is a representation vector.
After adding these two types information, we can obtain the representations of entities to be predicted as
\begin{equation}
    \begin{aligned}
        \mathbf{E}_{h}^{T} &= \texttt{BERT}(\texttt{prompt}_N(r,t)), \\
        \mathbf{E}_{t}^{T} &= \texttt{BERT}(\texttt{prompt}_N(h,r)),
    \end{aligned}
\end{equation}
and we can get the prediction logits $\mathbf{P}_h^T$ and $\mathbf{P}_t^T$ in the same way as Eq.~(\ref{eq:triple_predict}).
The overall loss of \lmencoder is defined as follows:
\begin{equation}
\label{eq:loss_text}
\resizebox{.89\columnwidth}{!}{$
\mathcal{L}_{\mathrm{text}}=\sum_{(h,r,t)\in \mathcal{T}}\big(\texttt{CrossEntropy}(\mathbf{P}_h^T, \mathbf{L}_{h}) + \texttt{CrossEntropy}(\mathbf{P}_t^T, \mathbf{L}_{t})\big),
$}
\end{equation}
where $\mathbf{P}_h$ and $\mathbf{L}_h$ are the prediction logits and labels for predicting the head entity, $\mathbf{P}_t$ and $\mathbf{L}_t$ are the prediction logits and labels for predicting the tail entity, respectively, given the triple $(h,r,t)$.

\subsection{Co-distillation Learning}
Considering the complementarity between \kgeencoder and \lmencoder, we propose to transfer the knowledge mutually by knowledge distillation, which is known as co-distillation.
The loss for traditional knowledge distillation with a fixed teacher model and a student model to be optimized is a combination of the classification loss and the Kullback Leibler (KL) divergence loss, defined as follows:
\begin{equation}
    \mathcal{L}_{s} = \texttt{CrossEntropy}(\mathbf{P}_s, \mathbf{L}) + \texttt{KL}(\mathbf{P}_t \,||\, \mathbf{P}_s),
\end{equation}
where $\mathcal{L}_s$ denotes the knowledge distillation loss for the student model. 
$\mathbf{P}_t$ and $\mathbf{P}_s$ denote the probabilities of all classes predicted by the teacher model and the student model, respectively.
$\mathbf{L}$ denotes the label vectors for a given instance.

For our \kgeencoder and \lmencoder, we empirically find that the scores for target entities both become higher in the training process.
It is reasonable because both of our models are trained to predict the target entities as top-1.
However, the scores for non-target entities are varying obviously.
Given that \kgeencoder is a structure-based model while \lmencoder is a PLM-based model, 
we think that the scores for non-target entities are also important to quantize the knowledge of models.
Therefore, the key point for co-distillation is to transfer the predicted probabilities of non-target entities mutually, rather than only to deliver the probabilities of target entities.

Recently, the importance of non-target classes has drawn more attention.
DKD \cite{DKD} decouples the classical knowledge distillation loss into two parts to release the potential of knowledge contained in non-target classes.
Motivated by this, we also emphasize the importance of non-target entities for KG embedding.
However, unlike the scenario for classical knowledge distillation with a fixed teacher, the knowledge is not guaranteed for co-distillation because a model can be the teacher and the student at the same time.
Hence, we designed a heuristic method for selective knowledge transfer.

In the rest of this section, we take the task of predicting tail entities as an example.
For an incomplete triplet $(h,r,\text{[MASK]})$, we obtain the prediction logits $\mathbf{P}_t^S$ and $\mathbf{P}_t^T$ from \kgeencoder and \lmencoder, respectively.
When \kgeencoder is the teacher model, we rank all entities in descending order according to $\mathbf{P}_t^S$ and select half entities with higher scores.
Then, we select the logits of these entities from $\mathbf{P}_t^{S}$ and $\mathbf{P}_t^{T}$ which are denoted by $\mathbf{P}_t^{S_1}$ and $\mathbf{P}_t^{T_1}$, respectively. 
The decoupled loss for $\mathbf{P}_t^{S_1}$ and $\mathbf{P}_t^{T_1}$ is
\begin{equation}
    \mathcal{L}_{KD}(\mathbf{P}_t^{S_1}, \mathbf{P}_t^{T_1}) = \texttt{KL}(\mathbf{b}_t^{S_1}\,||\,\mathbf{b}_t^{T_1}) + \texttt{KL}(\hat{\mathbf{P}}_t^{S_1}\,||\,\hat{\mathbf{P}}_t^{T_1}),
\end{equation}
where $\mathbf{b}_t^{S_1}=[p_1, 1-p_1] \in \mathbb{R}^{1 \times 2},\mathbf{b}_t^{T_1}=[p_2, 1-p_2] \in \mathbb{R}^{1 \times 2}$ denote the binary probabilities in terms of target entities, assuming that $p_1$ and $p_2$ are the probabilities for target entities from \kgeencoder and \lmencoder, respectively. 
$\hat{P}_t^{S_1}$ and $\hat{P}_t^{T_1}$ denote the probabilities excluded target entities from $P_t^{S_1}$ and $P_t^{T_1}$, respectively.
When \lmencoder is the teacher model, we can also get $\mathbf{P}_t^{S_2}$ and $\mathbf{P}_t^{T_2}$ in the same way.

\subsection{Put It All Together}
The proposed approach \modelname uses co-distillation learning to interact with \kgeencoder and \lmencoder for bidirectional knowledge transfer.
The learning losses for \kgeencoder and \lmencoder are
\begin{equation}
\label{eq:loss_distill}
\resizebox{.89\columnwidth}{!}{$
    \begin{aligned}
        \mathcal{L}_\text{\lmencoder} &= \alpha\, \mathcal{L}_{KD}(\mathbf{P}_t^{S_1}, \mathbf{P}_t^{T_1})
        + (1-\alpha)\, \texttt{CrossEntropy}(\mathbf{P}_t^T, \mathbf{L}_t), \\
        \mathcal{L}_\text{\kgeencoder} &= \beta\, \mathcal{L}_{KD}(\mathbf{P}_t^{T_2}, \mathbf{P}_t^{S_2})
        + (1-\beta)\, \texttt{CrossEntropy}(\mathbf{P}_t^S, \mathbf{L}_t),
    \end{aligned}
$}
\end{equation}
where $\mathbf{L}_t$ denotes the label vectors for predicting $t$. $\alpha$ and $\beta$ are hyper-parameters for balance.
\kgeencoder and \lmencoder are optimized jointly.
For each mini-batch, losses are computed from the same data and the two models are updated depending on their losses, respectively.
The training process is presented in Algorithm~\ref{alg:Train}.
In the inference process, given an incomplete triplet, we combine the prediction probabilities of \kgeencoder and \lmencoder as the final output probabilities by weighted averaging to rank candidates.

\begin{algorithm}[!tb]
\caption{Training process of \modelname.}
\label{alg:Train}
\KwIn{Training triplet set $\mathcal{T}_{\text{train}}$,
validation triplet set $\mathcal{T}_{\text{valid}}$,
test triplet set $\mathcal{T}_{\text{test}}$,
entity names and descriptions.}
\KwOut{Candidate ranking list for each incomplete triplet.}
Initialize model parameters and input embeddings\; 
Generate masked triplets and prompts from $\mathcal{T}_{\text{train}}$\; 
\For{$epoch$ $\leftarrow 1$ \KwTo $max\_epoch\_num$}{
    \For{$step \leftarrow 1$ \KwTo $max\_step\_num$}{
        $b\leftarrow$ sample a training batch\;
            \For{$(h, r, \text{[MASK]})$ in $b$}{
            Compute the logits and loss Eq.~(\ref{eq:loss_structure}) of \kgeencoder\;
            Compute the logits and loss Eq.~(\ref{eq:loss_text}) of \lmencoder\;
            Compute the losses Eq.~(\ref{eq:loss_distill}) of \modelname\;
        }
    Get the overall loss and update models\;
    }
    Validate \modelname using $\mathcal{T}_{\text{valid}}$\;
    \lIf{early stop}{break}
}
Test \modelname using $\mathcal{T}_{\text{test}}$\;
\end{algorithm}
\section{Experiments}\label{sect:exp}
In this section, we report the experimental results of the proposed approach \modelname.
The source code is available from GitHub.\footnote{\url{https://github.com/nju-websoft/CoLE}}

\subsection{Experiment Setup}

\subsubsection{Datasets}
We use two benchmark datasets FB15K-237 \cite{FB15K237} and WN18RR \cite{ConvE} to train and evaluate our approach. 
The two datasets remove some inverse edges from their previous versions (i.e., FB15K and WN18 \cite{TransE}) to prevent the leakage of test triplets into the training process.



\subsubsection{Evaluation Protocol}
For each triplet $(h,r,t)$ in the test set, 
we obtain two incomplete triplets, $(h,r,\text{[MASK]})$ and $(t,r^-,\text{[MASK]})$, for tail entity prediction and head entity prediction, respectively.
$r^-$ denotes the reverse relation for $r$.
The queries are fed into \kgeencoder to get output representations.
Then the prediction logits are calculated by an inner product between the output representations and all entity embeddings. 
We finally obtain the ranking lists for candidate entities by sorting the logits in descending order.
For \lmencoder or \modelname, the ranking lists are obtained in the same way.
We employ Hits@$k$ ($k = 1, 3, 10$) and MRR under the filtered setting \cite{TransE} to assess the performance.
Averaged results on head entity prediction and tail entity prediction are reported.

\subsubsection{Implementation Details}
We implement our method with PyTorch and all experiments are conducted on a workstation with two Intel Xeon Gold 6326 CPUs, 512GB memory and a NVIDIA RTX A6000 GPU.
We leverage the BERT-base model\footnote{Downloaded from \url{https://huggingface.co/}.} as the PLM for \lmencoder, 
and a vanilla Transformer encoder \cite{Transformer} for \kgeencoder.
We employ the AdamW optimizer and a cosine decay scheduler with linear warm-up for optimization.
We determine the hyper-parameter values by using the grid search based on the MRR performance on the validation set.
We select the layer number of Transformer in $\{1,2,4,8,12,16\}$,
the head number in $\{1,2,4,8\}$,
the batch size in $\{1024,2048,4096,8192\}$,
the learning rate for \kgeencoder in $\{ \text{1e-4},\text{2e-4},\text{3e-4},\text{5e-4}, \text{1e-3}\}$,
the learning rate for \lmencoder in $\{ \text{1e-5},\text{3e-5},\text{5e-5}, \text{1e-4}\}$,
the learning rate for co-distillation in $\{ \text{1e-5}, \text{3e-5},\text{5e-5}, \text{1e-4}\}$,
the values of $\alpha$ and $\beta$ in $\{0.1,0.2,\dots,0.9\}$.
Table~\ref{tab:best_hp} lists the picked values for important hyper-parameters.

\begin{table}[!t]
\centering
\caption{Selected hyper-parameter values for \modelname.}
{\small
\begin{tabular}{lccccc}
  \toprule
   & Dim. & \# Layers & \# Heads & $\alpha$ & $\beta$\\
  \midrule
  FB15K-237 & 256 & 8 & 2 & 0.5 & 0.8 \\ 
  WN18RR & 256 & 12 & 4 & 0.5 & 0.7 \\
  \bottomrule
\end{tabular}
}
\label{tab:best_hp}
\end{table}

\subsection{Baselines}

\begin{table*}[!t]
\caption{Link prediction results compared with structure-based baselines.}
\label{tab:kge_results}
\centering
{\small
    \begin{tabular}{lccccccccc}
        \toprule
        \multirow{3}{*}{Model} & \multicolumn{4}{c}{FB15K-237} && \multicolumn{4}{c}{WN18RR} \\
        \cmidrule{2-5} \cmidrule{7-10}
        & Hits@1 & Hits@3 & Hits@10 & MRR & & Hits@1 & Hits@3 & Hits@10 & MRR \\
        \midrule
        TransE & 0.231 & 0.367 & 0.528 & 0.329 && 0.013 & 0.400 & 0.528 & 0.223 \\
        ConvE & 0.237 & 0.356 & 0.501 & 0.325 && 0.400 & 0.440 & 0.520 & 0.430 \\
        RotatE & 0.241 & 0.375 & 0.533 & 0.338 && 0.428 & 0.492 & 0.571 & 0.476 \\
        TuckER & 0.266 & 0.394 & 0.544 & 0.358 && 0.443 & 0.482 & 0.526 & 0.470 \\
        CoKE & 0.272 & 0.400 & 0.549 & 0.364 && \underline{0.450} & 0.496 & 0.553 & 0.484 \\
        CompGCN & 0.264 & 0.390 & 0.535 & 0.355 && 0.443 & 0.494 & 0.546 & 0.479 \\
        ATTH & 0.252 & 0.384 & 0.540 & 0.348 && 0.443 & 0.499 & 0.573 & 0.486 \\
        DualE & 0.237 & 0.363 & 0.518 & 0.330 && 0.440 & 0.500 & 0.561 & 0.482 \\
        ConEx & 0.271 & 0.403 & 0.555 & 0.366 && 0.448 & 0.493 & 0.550 & 0.481 \\
        $\text{M}^2$GCN & 0.275 & 0.398 & \textbf{0.565} & 0.362 && 0.444 & 0.498 & 0.572 & 0.485 \\
        HittER & \textbf{0.279} & \underline{0.409} & \underline{0.558} & \textbf{0.373} && \textbf{0.462} & \textbf{0.516} & \textbf{0.584} & \textbf{0.503} \\
        \midrule
        \kgeencoder & \underline{0.277} & \textbf{0.412} & 0.556 & \underline{0.372} && 0.443 & 0.501 & 0.578 & 0.486 \\
        $\text{\kgeencoder}_{\text{co-distilled}}$ & \textbf{0.279} & \textbf{0.412} & 0.556 & \textbf{0.373} && 0.446 & \underline{0.504} & \underline{0.581} & \underline{0.489} \\
        \bottomrule
    \end{tabular}}
\end{table*}

\begin{table*}[!t]
\caption{Link prediction results compared with PLM-based baselines.}
\label{tab:lm_results}
\centering
{\small
    \begin{tabular}{lccccccccc}
        \toprule
        \multirow{3}{*}{Model} & \multicolumn{4}{c}{FB15K-237} && \multicolumn{4}{c}{WN18RR} \\
        \cmidrule{2-5} \cmidrule{7-10}
        & Hits@1 & Hits@3 & Hits@10 & MRR & & Hits@1 & Hits@3 & Hits@10 & MRR \\
        \midrule
        KG-BERT & - & - & 0.420 & - && 0.041 & 0.302 & 0.524 & 0.216 \\
        MLMLM & 0.187 & 0.282 & 0.403 & 0.259 && 0.440 & 0.542 & 0.611 & 0.502 \\
        kNN-KGE & 0.280 & 0.404 & 0.550 & 0.370 && 0.525 & 0.604 & 0.683 & 0.579 \\
        \midrule
        \lmencoder & \underline{0.287} & \underline{0.420} & \underline{0.562} & \underline{0.381} && \underline{0.529} & \underline{0.607} & \underline{0.686} & \underline{0.583} \\
        $\text{\lmencoder}_{\text{co-distilled}}$ & \textbf{0.293} & \textbf{0.426} & \textbf{0.570} & \textbf{0.387} && \textbf{0.532} & \textbf{0.607} & \textbf{0.689} & \textbf{0.585} \\
        \bottomrule
    \end{tabular}}
\end{table*}

\begin{itemize}
    \item Structure-based models.
    For structure-based KG embedding, we compare \kgeencoder against 11 representative link prediction models.
    We choose three geometric models, including TransE \cite{TransE}, RotatE \cite{RotatE} and DualE \cite{DualE}.
    We also choose the latest tensor decomposition model TuckER \cite{TuckER}.
    The left baselines are all based on deep neural networks.
    ConvE \cite{ConvE} and ConEx \cite{ConEx} use convolutions to capture the interactions between entities and relations.
    CompGCN \cite{CompGCN} and $\text{M}^2$GCN \cite{M2GCN} leverage GCNs.
    ATTH \cite{ATTH} is a hyperbolic embedding model to capture the hierarchical patterns in a KG.
    We also compare \kgeencoder with CoKE \cite{CoKE} and HittER \cite{hitter}, which are the most relevant studies based on the Transformer architecture.
    CoKE \cite{CoKE} only focuses on modeling a single triplet, while HittER \cite{hitter} introduces entity neighborhood as the contextual information but without entity reconstruction.
    
    \item PLM-based models.
    For PLM-based KG embedding, we pick 3 representative models as baselines, including KG-BERT \cite{kg-bert}, MLMLM \cite{mlmlm} and kNN-KGE \cite{knnkge}.
    KG-BERT is a triplet classification model and only utilizes entity names and relation names as text information.
    It can also do link prediction but is time-consuming.
    MLMLM and kNN-KGE further utilize the descriptions of entities as additional information, which are both link prediction models.
    MLMLM adds multiple [MASK] tokens to predict entities with multi-token names, and kNN-KGE learns initial representations for all entities from descriptions before training.
    
    \item Ensemble methods. 
    To our knowledge, no previous work considers knowledge transfer between structured-based and PLM-based link prediction.
    We design two ensemble methods, namely ProbsMax and ProbsAvg, in \modelname.
    ProbsMax selects the maximum probabilities of \kgeencoder and \lmencoder for a given entity to rank.
    ProbAvg averages the prediction probabilities.
    The two variants of \modelname are denoted by $\text{ProbsMax}_\text{co-distilled}$ and $\text{ProbsAvg}_\text{co-distilled}$, respectively.
\end{itemize}

The results of all baselines are taken from their original papers, with the exception of TransE, which appeared earlier than the two datasets.
We reproduce its results using OpenKE \cite{OpenKE}.


\begin{table*}
\caption{Link prediction results of ensemble methods.}
\label{tab:distilll_results}
\centering
{\small
    \begin{tabular}{lccccccccc}
        \toprule
        \multirow{3}{*}{Model} & \multicolumn{4}{c}{FB15K-237} && \multicolumn{4}{c}{WN18RR} \\
        \cmidrule{2-5} \cmidrule{7-10}
        & Hits@1 & Hits@3 & Hits@10 & MRR & & Hits@1 & Hits@3 & Hits@10 & MRR \\
        \midrule
        \kgeencoder & 0.277 & 0.412 & 0.556 & 0.372 && 0.443 & 0.501 & 0.578 & 0.486 \\
        \lmencoder & 0.287 & 0.420 & 0.562 & 0.381 && 0.529 & 0.607 & 0.686 & 0.583 \\
        ProbsMax & 0.291 & 0.427 & 0.570 & 0.386 && 0.517 & 0.600 & 0.685 & 0.574 \\
        ProbsAvg & \textbf{0.294} & \textbf{0.430} & \textbf{0.574} & \textbf{0.389} && \underline{0.530} & \textbf{0.609} & \underline{0.692} & \underline{0.585} \\
        \midrule
        $\text{\kgeencoder}_{\text{co-distilled}}$ & 0.279 & 0.412 & 0.556 & 0.373 && 0.446 & 0.504 & 0.581 & 0.489 \\
        $\text{\lmencoder}_{\text{co-distilled}}$ & \underline{0.293} & 0.426 & 0.570 & \underline{0.387} && \textbf{0.532} & 0.607 & 0.689 & \underline{0.585} \\
        $\text{ProbsMax}_\text{co-distilled}$ & 0.292 & \underline{0.428} & \underline{0.571} & \underline{0.387} && 0.518 & 0.595 & 0.684 & 0.574 \\
        $\text{ProbsAvg}_\text{co-distilled}$ & \textbf{0.294} & \textbf{0.430} & \textbf{0.574} & \textbf{0.389} && \textbf{0.532} & \underline{0.608} & \textbf{0.694} & \textbf{0.587} \\
        \bottomrule
    \end{tabular}
}
\end{table*}

\begin{table*}
\caption{Ablation results of \kgeencoder and \lmencoder.}
\label{tab:ablation}
\centering
{\small
    \begin{tabular}{lccccccccc}
        \toprule
        \multirow{3}{*}{Model} & \multicolumn{4}{c}{FB15K-237} && \multicolumn{4}{c}{WN18RR} \\
        \cmidrule{2-5} \cmidrule{7-10}
        & Hits@1 & Hits@3 & Hits@10 & MRR & & Hits@1 & Hits@3 & Hits@10 & MRR \\
        \midrule
        $\text{\kgeencoder}_{\text{w/o neighbor}}$ & 0.269 & 0.399 & 0.548 & 0.362 && 0.437 & 0.478 & 0.523 & 0.466 \\
        \kgeencoder & 0.277 & 0.412 & 0.556 & 0.372 && 0.443 & 0.501 & 0.578 & 0.486 \\
        \midrule
        $\text{\lmencoder}_{\text{w/o soft prompt}}$ & 0.264 & 0.389 & 0.527 & 0.354 && 0.510 & 0.579 & 0.672 & 0.563 \\
        $\text{\lmencoder}_{\text{w/o description}}$ & 0.271 & 0.408 & 0.556 & 0.368 && 0.414 & 0.488 & 0.571 & 0.467 \\
        $\text{\lmencoder}_{\text{w/o neighbor}}$ & 0.283 & 0.418 & 0.561 & 0.378 && 0.514 & 0.598 & 0.685 & 0.573 \\
        \lmencoder & 0.287 & 0.420 & 0.562 & 0.381 && 0.529 & 0.607 & 0.688 & 0.582 \\
        \bottomrule
    \end{tabular}
}
\end{table*}

\subsection{Results and Analyses}
Here we report and analyze the main results.

\smallskip\noindent\textbf{Results of structure-based models.}
Experimental results for structure-based models are shown in Table \ref{tab:kge_results}.
It takes $7.9$ hours and $6.2$ hours to train \kgeencoder with the best parameters on FB15k-237 and WN18RR, respectively.
We can observe that \kgeencoder achieves competitive performance on the two datasets.
On FB15K-237, the performance of \kgeencoder is nearly close to that of the state-of-the-art model HittER.
Both models are based on the Transformer architecture and leverage neighborhood information.
They outperform CoKE stably which neglects such context.
On WN18RR, our model does not perform best.
It is slightly inferior to HittER and comparable with other competitive methods.
We think this is because WN18RR has more hierarchical structures and HittER benefits from its hierarchical architecture to capture such patterns.
The characteristics of the dataset are also beneficial for ATTH and DualE, despite their underwhelming performance on FB15K-237.
ATTH leverages several hyperbolic operations to distinguish the hierarchies.
DualE can capture complicated relations with unified operations, including translation and rotation.
We also notice that \kgeencoder gets further improvement after co-distillation.
It further indicates the complementarity between \kgeencoder and \lmencoder, and co-distillation makes them benefit from each other.
We give more analyses shortly in Section~\ref{sect:further_analyses}.

\smallskip\noindent\textbf{Results of PLM-based models.}
The results for PLM-based models are shown in Table \ref{tab:lm_results}.
It takes $8.8$ hours and $3.2$ hours to train \lmencoder with the best parameters on FB15k-237 and WN18RR, respectively.
We can see that \lmencoder even without distillation outperforms other baselines on both datasets.
Among PLM-based models, KG-BERT does not work well since its prompt sentences do not give enough evidence, showing that additional information is important for PLM-based models.
kNN-KGE and MLMLM use the descriptions of entities for link prediction and they obtain a significant improvement.
However, the extent of improvement differs due to their utilization of the descriptions of entities.
In addition to adding the descriptions as support information into the prompt sentences, kNN-KGE further learns new representations of entities based on the descriptions before the model training.
Thanks to our soft prompts and neighbor information in prompt sentences, \lmencoder outperforms kNN-KGE stably.
We also notice that \lmencoder obtains an improvement with co-distillation, similar to \kgeencoder.

\smallskip\noindent\textbf{Results of ensemble methods.}
Table~\ref{tab:distilll_results} lists the results of ensemble methods.
It takes $8.7$ hours to co-distill \kgeencoder and \lmencoder on FB15k-237 and $1.4$ hours on WN18RR, with the best parameters.
We notice that both ensemble methods achieve a further improvement no matter whether we leverage co-distillation.
The improvement is in accord with the observation found by existing work \cite{ConEx} that ensemble learning with a simple combination can increase the performance.
For example, the baseline ensemble methods ProbsMax and ProbsAvg can improve both \kgeencoder and \lmencoder on FB15K-237.
ProbsAvg performs better than ProbsMax.
The maximum probabilities emphasize some wrong predictions of the two models, making ProbsMax unstable.
Furthermore, the performance of our model variants is slightly better than the ensemble baselines without co-distillation.
This is in line with our intuition that ensemble learning performs better with stronger base models.
As co-distillation mainly transfers mutual knowledge without generating much new knowledge, such improvement is not significant.
We leave it as future work to explore other ensemble methods.

\subsection{Ablation Study}
The results are shown in Table \ref{tab:ablation}.

\smallskip\noindent\textbf{Ablation study for \kgeencoder.}
The only additional information that we utilize for \kgeencoder is neighboring triplets, so we conduct an ablation study to verify the influence of neighbor information.
The neighbor information significantly improves \kgeencoder on both datasets in terms of Hits@1, 0.269 $\rightarrow$ 0.277 on FB15K-237 and 0.437 $\rightarrow$ 0.443 on WN18RR.
This further indicates the effectiveness of the entity neighborhood in prediction, and our proposed \kgeencoder can capture such contextual information well.

\smallskip\noindent\textbf{Ablation study for \lmencoder.}
We apply three kinds of support textual information to construct the prompts for \lmencoder, namely \textit{soft prompt}, \textit{entity description} and \textit{neighbor information}.
We hereby conduct an ablation study on the support information to verify their effectiveness.
As we can see, all of them contribute a lot to \lmencoder, and the contributions are varied for different datasets.
We argue that the effectiveness of different kinds of textual information is related to the graph structure of KGs.
FB15k-237 is a much more dense KG with fewer entities and more relations.
Soft prompts can better enhance the expression of relations between entities in this dataset.
For WN18RR, entity descriptions play a more important role.
This is because that WN18RR has a large number of entities (about 41K).
The additional description information can work more obviously to distinguish these entities.
As for the neighbor information, the Hits@1 improvement is more significant on WN18RR than that on FB15K-237 (0.514 $\rightarrow$ 0.529 vs 0.283 $\rightarrow$ 0.287).
Given that we only sample a few neighboring triplets, neighbor information contributes more for a more sparse KG.
Overall, we demonstrate the effectiveness of all the proposed support information.

\subsection{Further Analyses}
\label{sect:further_analyses}
\begin{figure}[!t]
\centering
\includegraphics[width=\linewidth]{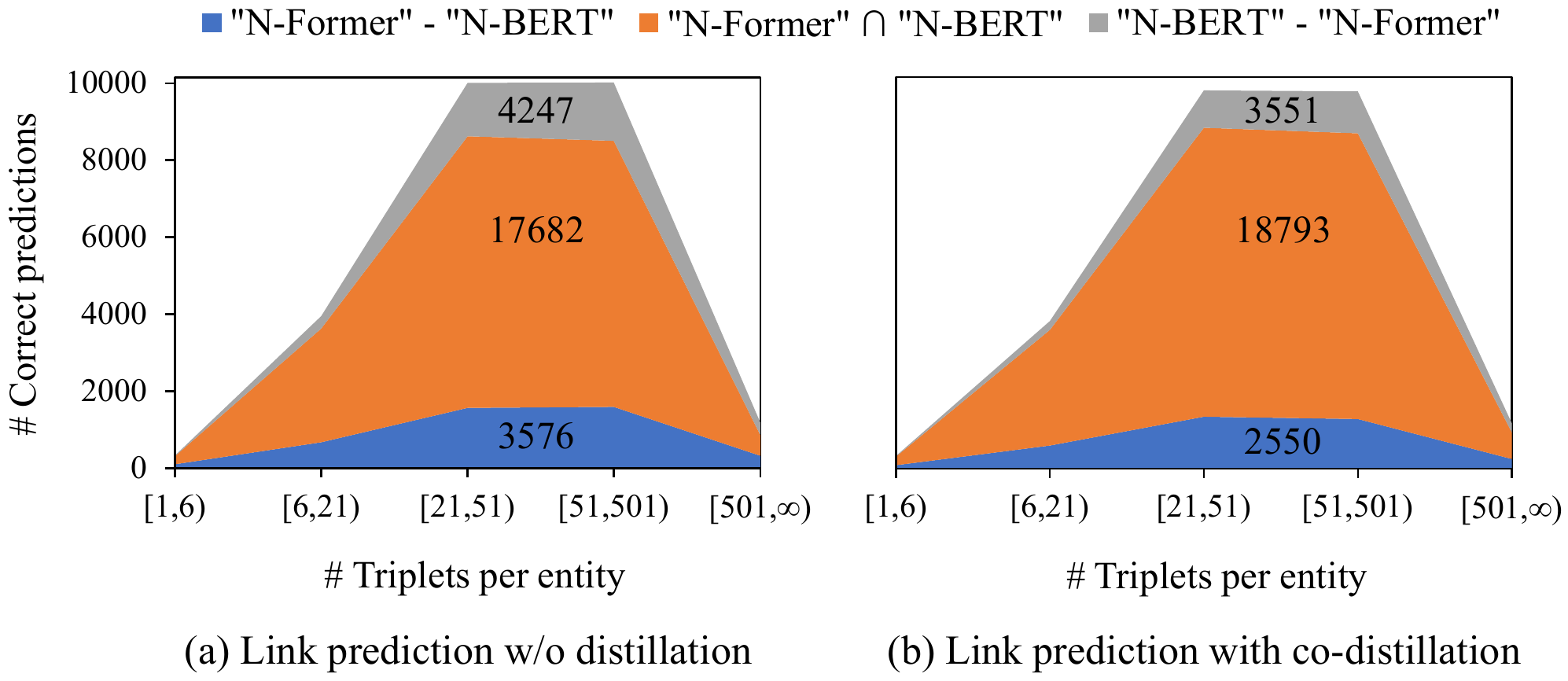}
\caption{Correct predictions of \kgeencoder and \lmencoder when using no-distillation and co-distillation on FB15K-237.
The blue area denotes the right triplets predicted by \kgeencoder which exclude those that can also be predicted by \lmencoder.
The orange area denotes the overlap triplets predicted correctly by \kgeencoder and \lmencoder. 
The gray area denotes the right triplets predicted by \lmencoder which exclude those that can also be predicted by \kgeencoder.
}
\label{fig:complement_before_after_distill}
\vspace{5pt}
\end{figure}

\noindent\textbf{Complementarity in our model.}
We check the correct entity prediction results of \kgeencoder and \lmencoder in overlap on FB15K-237.
We also compare the results with and without co-distillation, as shown in Figure~\ref{fig:complement_before_after_distill}.
When using no-distillation, 3576 triplets can be predicted correctly only by \kgeencoder.
For \lmencoder, there are 4247 correct triplets.
The blue and gray areas represent the two kinds of triplets. 
These triplets reflect the complementarity of \kgeencoder and \lmencoder.
We can observe that the two areas get shrunk (i.e., the complementarity becomes weaker) if we leverage the co-distillation.
In turn, the orange area, which denotes the overlap of the predicted triplets, expands.
It indicates that co-distillation indeed leverages the complementarity and transfers the unique knowledge of each model mutually, thus benefiting them.

\begin{figure}[!t]
\centering
\includegraphics[width=\linewidth]{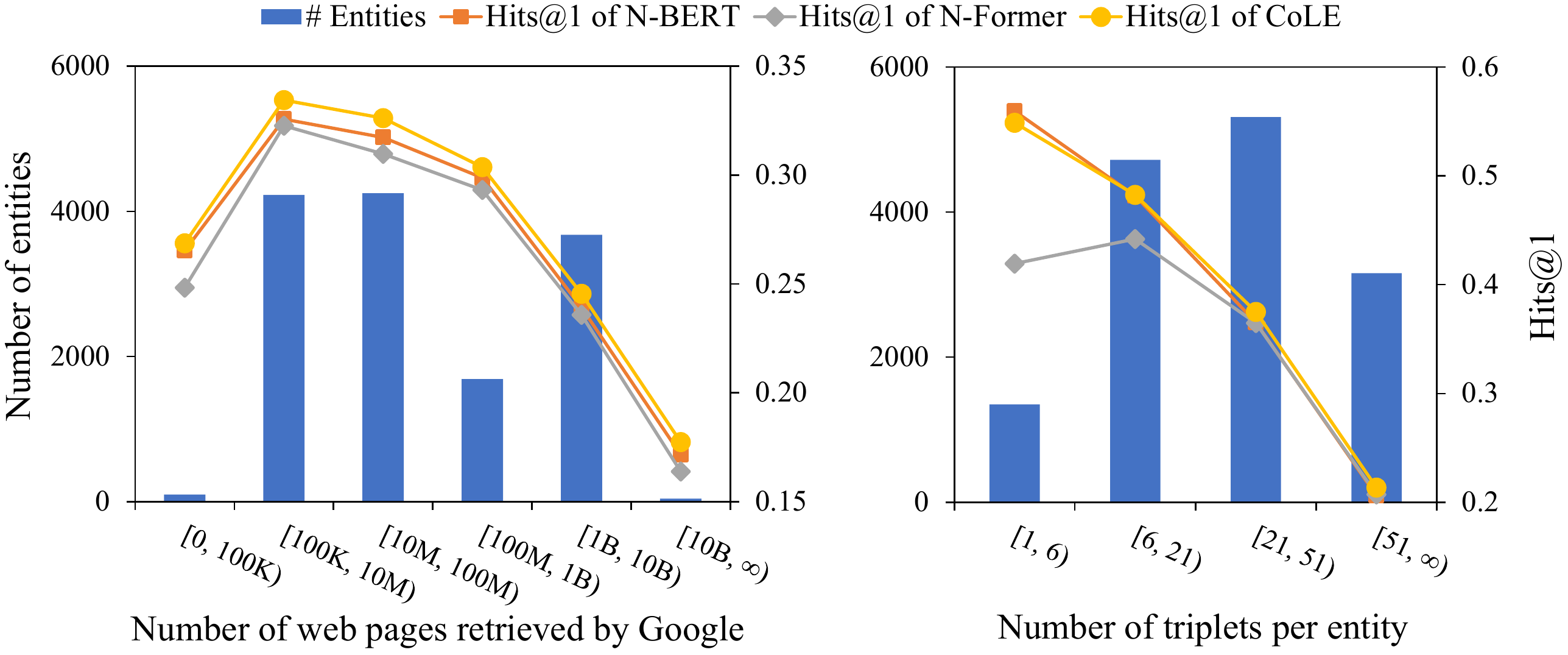}
\caption{
Hits@1 results on FB15K-237 along with the numbers of retrieved web pages and edges per entity, resp.
}
\label{fig:long_tail_ours}
\end{figure}

\smallskip\noindent\textbf{Result analyses for long-tail and popular entities.}
Following the practices in Section~\ref{chapter:intro}, we partition the entities of FB15K-237 into several groups based on the number of retrieved web pages and the number of edges per entity, respectively.
When partitioning by the number of web pages, \kgeencoder and \lmencoder have the same tendency. 
Their performance both increases at first and then decreases along with the growing number of web pages.
Moreover, \modelname outperforms them all the time.
As the number of web pages cannot fully indicate whether an entity is popular or long-tail, the complementarity between \kgeencoder and \lmencoder is not obvious.
We further partition entities according to the number of triplets.
\lmencoder surpasses \kgeencoder by a large margin for long-tail entities, 
while \kgeencoder is slightly better for popular entities.
Although the scores of \lmencoder and \kgeencoder on popular entities are close, the complementarity cannot be omitted because the number of triplets related to popular entities is large.
There are still many triplets that can be correctly predicted only by \kgeencoder.
Overall, the analyses above verify the complementarity between \kgeencoder and \lmencoder on both popular and long-tail entities.

\begin{figure}[!t]
\centering
\includegraphics[width=0.85\linewidth]{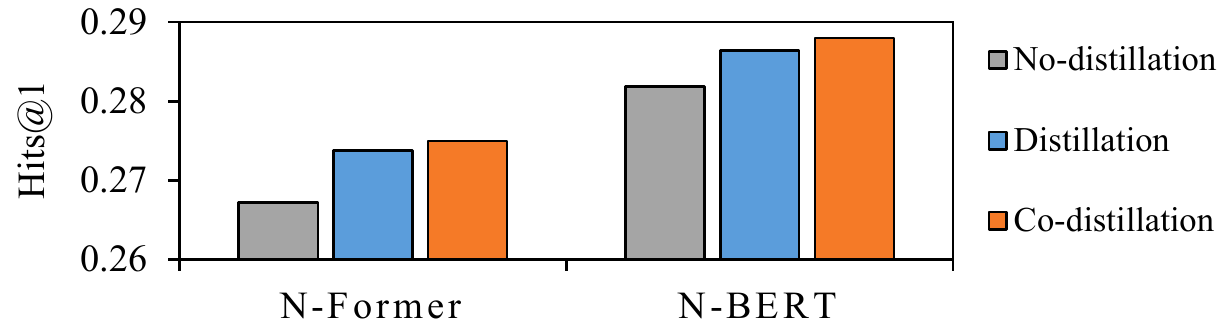}
\caption{Hits@1 results of our models when using no-distillation, the conventional (unidirectional) distillation and our co-distillation on FB15K-237.}
\label{fig:distillation}
\end{figure}

\smallskip\noindent\textbf{Comparison of different distillation methods.}
We also explore the effectiveness of different distillation methods.
We choose the conventional (unidirectional) distillation for comparison.
In this method, we fix the teacher model and only train the student model through the distillation.
Figure~\ref{fig:distillation} shows the results of \kgeencoder and \lmencoder without leveraging neighborhood information on FB15K-237.
We observe that both methods can obtain an improvement, as distillation can generally introduce new knowledge for models.
Moreover, co-distillation outperforms unidirectional distillation stably.
We believe this is because that the co-distillation adopts a mutual learning strategy and allows both models to learn from each other, while the unidirectional distillation neglects such interactions.
This further verifies the superiority of our co-distillation.

\section{Conclusion and Future Work}
In this paper, we present a co-distillation learning method that seeks effective knowledge transfer and mutual enhancement between structure-based and PLM-based KG embedding models.
For structure-based KG embedding, we propose \kgeencoder that reconstructs and predicts the missing entity of an incomplete triplet based on its relational neighbors.
For PLM-based KG embedding, we propose \lmencoder that generates the missing entity representation by probing BERT with a prompt of entity names, descriptions, and neighbors.
Our co-distillation learning method \modelname first decouples the prediction logits of the two models and then lets them teach their useful knowledge to each other by bidirectional knowledge transfer with logit distillation.
Experiments on FB15K-237 and WN18RR show that \kgeencoder and \lmencoder achieve competitive and even the best results compared with existing work.
The ensemble method \modelname advances the state-of-the-art of KG embedding.

In future work, we plan to investigate knowledge transfer between multi-source KGs and experiment with additional KG embedding tasks such as entity alignment.

\smallskip\noindent\textbf{Acknowledgments.} 
This work was supported by National Natural Science Foundation of China (No. 62272219) and Beijing Academy of Artificial Intelligence (BAAI).


\balance
\bibliographystyle{ACM-Reference-Format}
\bibliography{reference}

\end{document}